\documentclass[%
 reprint,
 superscriptaddress,
nofootinbib,
 amsmath,amssymb,
 aip,
]{revtex4-2}

\usepackage{graphicx}% Include figure files
\usepackage{dcolumn}% Align table columns on decimal point
\usepackage{bm}% bold math
\usepackage{hyperref}% add hypertext capabilities
\usepackage{physics}
\usepackage{float}
\usepackage{xcolor}
\usepackage{comment}

\usepackage{tabularray}

\usepackage{balance}

\begin{document}

%\preprint{APS/123-QED}

\title{LLMs Can Assist with Proposal Selection at Large User Facilities}% Force line breaks with \\
\author{Lijie Ding}
\email{dingl1@ornl.gov}
\affiliation{Neutron Scattering Division, Oak Ridge National Laboratory, Oak Ridge, TN 37831, USA}

\author{Janell Thomson}
\affiliation{Neutron Scattering Division, Oak Ridge National Laboratory, Oak Ridge, TN 37831, USA}

\author{Jon Taylor}
\affiliation{Neutron Scattering Division, Oak Ridge National Laboratory, Oak Ridge, TN 37831, USA}

\author{Changwoo Do}
\email{doc1@ornl.gov}
\affiliation{Neutron Scattering Division, Oak Ridge National Laboratory, Oak Ridge, TN 37831, USA}

\date{\today}% It is always \today, today,

\begin{abstract}
We explore how large language models (LLMs) can enhance the proposal selection process at large user facilities, offering a scalable, consistent, and cost-effective alternative to traditional human review. Proposal selection depends on assessing the relative strength among submitted proposals; however, traditional human scoring often suffers from weak inter-proposal correlations and is subject to reviewer bias and inconsistency. A pairwise preference-based approach is logically superior, providing a more rigorous and internally consistent basis for ranking, but its quadratic workload makes it impractical for human reviewers. We address this limitation using LLMs. Leveraging the uniquely well-curated proposals and publication records from three beamlines at the Spallation Neutron Source (SNS), Oak Ridge National Laboratory (ORNL), we show that the LLM rankings correlate strongly with the human rankings (Spearman $\rho\simeq 0.2-0.8$, improving to $\geq 0.5$ after 10\% outlier removal). Moreover, LLM performance is no worse than that of human reviewers in identifying proposals with high publication potential, while costing over two orders of magnitude less. Beyond ranking, LLMs enable advanced analyses that are challenging for humans, such as quantitative assessment of proposal similarity via embedding models, which provides information crucial for review committees.
\end{abstract}

\maketitle

%\tableofcontents

\section{Introduction}

% 1. talk about proposal review, background and challenge, threw in some stats for SNS
Large user facilities, such as the Spallation Neutron Source\cite{alonso1999spallation,henderson2014spallation} (SNS) at Oak Ridge National Laboratory (ORNL), play a critical role in advancing scientific research by providing access to specialized instruments for experiments in fields like materials science\cite{mason2006spallation,melnichenko2007small}, chemistry\cite{trouw1999chemical}, and physics\cite{lindner2024neutrons,furrer2009neutron,beckurts2013neutron}. For each run cycle, typically occurring twice a year, SNS receives more than 500--600 proposals from researchers worldwide, requesting beam time to conduct experiments.\footnote{\url{https://neutrons.ornl.gov/sns}} These proposals fall into several categories, of which the most common are general user proposals. Traditionally, the proposal selection process is based on individual scoring carried out by human experts, as shown in Fig.~\ref{fig:approaches_illustration}. It faces several significant challenges. Proposals are distributed among a panel of reviewers (often domain-specific scientists), with each proposal typically evaluated by no more than 3 experts and each expert handling up to 9 proposals per cycle at SNS. Reviewers assign individual scores based on criteria like scientific merit, feasibility, and potential impact. These individual scores are then aggregated and used to rank the proposals for beam time approval. 

However, this approach often results in weak inter-proposal consistency, since reviewers assess subsets of proposals independently, comparisons across the full pool are indirect, leading to inconsistencies in scoring scales and relative rankings. Human factors exacerbate this: reviewer bias\cite{lin2023automated} (e.g., influenced by personal expertise or institutional affiliations), day-to-day variations in mood or fatigue, and cognitive biases from the order in which proposals are read can all skew evaluations. Moreover, the workload becomes burdensome with such large number of submitted proposals, contributing to reviewer burnout and higher costs in terms of time and labor. These issues not only compromise the fairness and reliability of selections but also limit the ability to perform advanced analyses, such as quantitative similarity checks between proposals to detect duplicates or overlaps.

% 2. talk about LLMs and some related work on article review, point out the fundamental difference between article and proposal review
Recent developments in Large language models (LLMs)\cite{zhao2023survey,xiao2025foundations,naveed2025comprehensive} show a promising direction for improving and automating proposal selection. Frontier LLMs such as GPT\cite{achiam2023gpt}, Gemini\cite{team2023gemini}, Claude, Grok, Qwen\cite{bai2023qwen}, and Kimi\cite{team2025kimi}, have demonstrated remarkable capabilities in natural language processing tasks\cite{jin2024agentreview,chollet2024arc,wang2024mmlu,zellers2019hellaswag}, including summarization, sentiment analysis, and content generation, making them suitable for evaluating complex scientific texts. As for scientific peer review, LLMs have been increasingly explored for automating manuscript assessments\cite{liang2024can, checco2021ai, zhuang2025large}, and it is no secret that many researchers use LLMs to assist with the review process. Nevertheless, proposal selection is fundamentally different from literature review. While these approaches in literature review focus on assessing the absolute quality of the work, proposal selection only cares about the relative strength between all proposals. Regardless of the overall quality of the submitted proposals, user facilities have to rank and select the top proposals due to limited resources. Therefore, the inter-proposal comparison is crucial for proposal selections.

% 3. what we are doing in this work.
In this work, we address these challenges and the gap in LLM use for proposal ranking by introducing a LLM-empowered pairwise preference approach for proposal selection. Instead of reviewing and scoring proposals individually, we use LLMs to evaluate every pair of proposals and determine preferences. Based on the win-lose results, we calculate the relative strength using the Bradley-Terry model. Applying this approach on historical beamline proposals, we evaluate the LLM rank against human rank by calculating the Spearman correlation for the past 20 run cycles. We also investigate the correlation between proposal ranking and associated publication record. We then carry out cost analysis to quantify the cost saving from the LLMs approach. Finally, we demonstrate that the embedding model from the LLMs can be used to analyze the similarity among proposals, which is a task human task struggle with.

\section{Method}

\subsection{Pairwise preference}

\begin{figure}[!t]
    \centering
    \includegraphics[width=\linewidth]{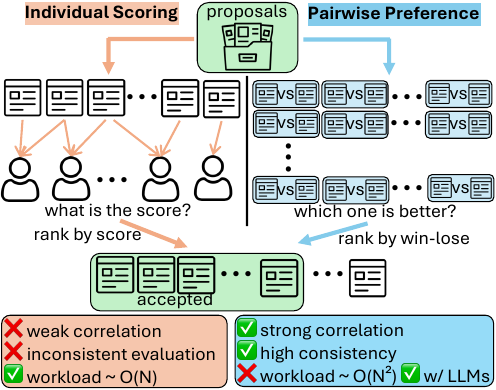}
    \caption{Two approaches for proposal selection. individual scoring (IS) aggregates independent scoring from different reviewers and rank the proposal based on the scores. pairwise preference (PP) compares all pairs of proposals and rank the proposals based on win-lose results. PP scales the workload quadratically, but can be handled with LLMs.}
    \label{fig:approaches_illustration}
\end{figure}

Proposal selection inherently focuses on relative strengths among submissions, so we adopt a pairwise preference\cite{furnkranz2003pairwise} (PP) approach to rank the proposals based on win-lose preference results among all pairs of proposals. The PP approach is logically superior to the currently used individual scoring (IS) approach, and the only drawback, quadratically increased workload with increasing number of proposals, can be resolved by using LLMs.

Fig.~\ref{fig:approaches_illustration} shows a comparison between the IS and PP approach. In the IS, all of the proposals are distributed to different reviewers. For SNS, each proposal is reviewed by at most 3 human expert, and each human expert is assigned with up to 9 proposals. Due to the significant human-human difference, the proposals reviewed by different people are weakly correlated, leading to inconsistent evaluation. In addition, even for proposals reviewed by the same human reviewer, the day-to-day change of human mood and knowledge also cause inconsistency. As the human reviewers' memory are affected by the proposal they have seen, the order of these proposals been reviewed introduce cognition bias.

Unlike the IS, the PP, even done with human is fundamentally a better approach. In PP, the ranking task is broken down in to many preference task that comparing a pair of proposals, which is a simpler cognition task than giving absolute score\cite{clark2018rate,perez2019pairwise}. Because every proposal is compared with every other ones, PP provide a strong correlation and high consistency. For $N$ proposals, PP go through all $N(N-1)/2$ pairs, and the rank is decided based on all of these win-lose results. The issue for PP is that the workload for ranking proposal increase quadratically with $N$, which make this superior approach impractical for human experts to perform. Fortunately, we can use LLMs to do this.

\subsection{Large language models}
To handle the $O(N^2)$ scale of the workload introduced by the pairwise preference approach. we use LLMs\cite{naveed2025comprehensive} to judge which proposal is preferred between any two. As all of the proposals submitted are in PDF format, which is not LLM-friendly, we first convert all of the proposal PDF documents into markdown format using an Optical Character Recognition (OCR) model\cite{mithe2013optical,MistralAI_OCR}. We then sort all of the proposals into corresponding run cycles, and find the LLM-determined win-lose results for every pair of the proposals within that cycle using the following system prompt and user prompt.

\smallskip
\noindent
\fbox{%
\begin{minipage}{1\linewidth}
\small
\textbf{System prompt:} You are an expert scientific reviewer for the ORNL Neutron Sciences General User Program. Your role is to compare two proposals based on scientific merit, providing a numerical score and substantive, constructive comments. Scientific merit is the primary consideration. Assume the proposal has already passed initial feasibility review by instrument scientists.
\end{minipage}%
}
\smallskip

\noindent
\fbox{%
\begin{minipage}{1\linewidth}
\small
\textbf{User prompt:} Please evaluate and compare the following two proposals:

Proposal A: \{proposal\_text\_a\}

Proposal B: \{proposal\_text\_b\}

Respond only with valid JSON in this exact structure (no additional text outside the JSON):
\{\\
  "summary": "[Concise summary of each proposal's scientific goals and methods]", \\
  "comparison": "[summarize aspects Proposal A vs. Proposal B]", \\
  "reasoning": "[Detailed reasoning which is better and why, only decide the winner after thorough comparison]", \\
  "winner": ["A" or "B" or "Tie"],
\}
\end{minipage}%
}
\smallskip

\begin{figure*}[!t]
    \centering
    \includegraphics[width=\linewidth]{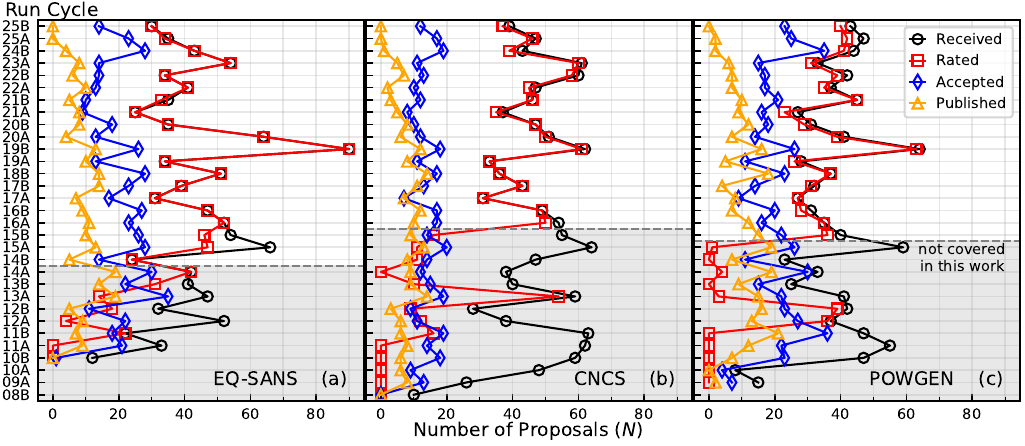}
    \caption{Number of proposals $N$ over the years, categorized by received, rated, accepted, and publication statue, for beamline (a ) EQ-SANS: Extended Q-Range Small-Angle Neutron Scattering Diffractometer, (b) CNCS: Cold Neutron Chopper Spectrometer, and (c) POWGEN: Powder Diffractometer. Earlier data are not covered in this work due to lack of rating from underdeveloped database. Data up-to-date as of Nov 10th, 2025. }
    \label{fig:proposal_stats}
\end{figure*}

The \{proposal\_text\_a\} and \{proposal\_text\_b\} part in the user prompt are replaced with the text of each proposal, respectively. In the user prompt, we instruct the LLMs to firstly summarize each proposal, then compare between two proposals, and do some reasoning about the comparison, and only decide the winner after all of these. In this way, all of the compute for deciding the winner are distributed across all of the output tokens, increase the reliability of the output. In addition, we also use the embedding model from the LLM to calculate the embedding vector of all proposals in the past 20 run cycles, which are use for similarity analysis in Sec.~\ref{ssec:LLMs_for_similarity_analysis}.

In this work, we use OpenRouter as the API provider all of the LLMs access. For data security consideration, we only use model provider with zero data retention. For the proposal pdf to markdown text conversion, we use Minstra OCR\cite{MistralAI_OCR} provided directly by OpenRouter. For pairwise preference judge, we use Gemini-2.5-flash\cite{comanici2025gemini} provided by Google Vertex. Finally, for the embedding calculation, we use Qwen3-embedding-8b\cite{zhang2025qwen3} (4096 dimension) provided by DeepInfra.

\subsection{Bradley-Terry model}
We use the Bradley-Terry (BT) model\cite{bradley1952rank} to calculate the relative strength from the pairwise preference results. Bradley-Terry model is a probabilistic framework for estimating latent strength parameters from head-to-head outcomes. For any two proposals $i$ and $j$, the model assumes the probability that $i$ is preferred over $j$ is given by:
\begin{equation}
    P(i > j) = \frac{s_i}{s_i + s_j}  
\end{equation}
where $s_i > 0$ and $s_j > 0$ represent the relative scientific merit or priority of proposals $i$ and $j$, respectively.

The BT score for each proposal $\{s_i\}$ are estimated via maximum likelihood using the minorization-maximization (MM) algorithm\cite{hunter2004mm}, an iterative fixed-point method that converges to the global optimum. After initializing scores uniformly, the score for each proposal are updated based on the interaction rule:
\begin{equation}
  s_i^{(k+1)} = \frac{w_i}{\sum_{j \neq i} \frac{n_{ij}}{s_i^{(k)} + s_j^{(k)}}}  
\end{equation}
where $w_i$ is the total number of wins (including 0.5 for ties) achieved by proposal $i$,  $n_{ij}$ is the number of comparisons between $i$ and $j$, $k$ is the iteration. Ties were treated as half-wins to each proposal. Scores are normalized at each step to sum to 1 to resolve scale invariance. The convergence was determined when the maximum relative change in any score fell below $10^{-8}$, or reaching maximum number of iteration 10000.

\section{Results}
\label{sec:Results}
We start by reviewing the statistics of the three most representative beamlines and investigating the correlation between LLM-enabled pairwise preference ranking and historical human individual scoring. Then, we evaluate both against publication record, analyze cost, and demonstrate embedding-based similarity analysis.

\subsection{Proposal statistics}
Among SNS's 20 beamlines, we focused on the three representative ones: the Extended Q-Range Small-Angle Neutron Scattering Diffractometer (EQ-SANS)\cite{zhao2010extended,liu2011first} for small-angle neutron scattering, the Cold Neutron Chopper Spectrometer (CNCS)\cite{ehlers2011new,ehlers2016cold} for inelastic scattering, and the Powder Diffractometer (POWGEN)\cite{huq2011powgen,huq2019powgen} for  powder diffraction. EQ-SANS, for example, received about 30 to 50 general user proposals each cycle, as high as 90, with acceptance rate ranging from $20\%-65\%$ due to limited instrument availability. The acceptance of these proposals are determined by experts in the field through the external peer review process. We will only consider this type of proposal in this work.  Fig.~\ref{fig:proposal_stats} illustrates the statistics by received, rated, accepted, and publication status. Earlier data are excluded due to incomplete ratings from the database. The proposal selection process is essential to ensure that the most scientifically meritorious projects are prioritized, maximizing the facility's impact on scientific discoveries and publications.

\subsection{Ranking correlation}
\label{ssec:Ranking_correlation}

\begin{figure}[!t]
    \centering
    \includegraphics[width=\linewidth]{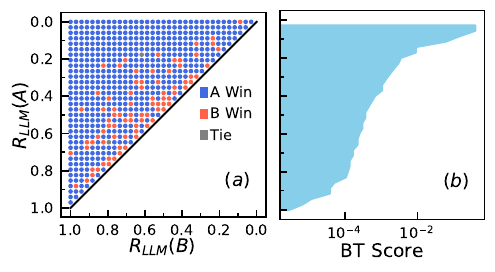}
    \caption{Using pairwise preference (PP) to rank proposals. (a) Example of the PP result for all pair of rated proposals from the EQ-SANS 20B run cycle. (b) Bradley-Terry (BT) score calculated based on the win-lose results.}
    \label{fig:llm_comparison}
\end{figure}

By gathering the win-lose results for all pairs of proposals within each cycle, we are able to calculate the BT score for each proposal. Fig.~\ref{fig:llm_comparison}(a) shows the win-lose results for the run cycle 20B for EQ-SANS, with the proposals arranged by the normalized rank $R_{LLM} = (\textit{rank}-1)/(N-1)$ based on the BT score. As expected, the higher ranked proposal (smaller $R_{LLM}$) more likely to win against the lower ranked one, thus the plot is mostly blue (Win). Fig.~\ref{fig:llm_comparison}(b) shows the BT score for each proposal, with monotonic decaying as the proposal rank drop.

\begin{figure*}[!th]
    \centering
    \includegraphics[width=\linewidth]{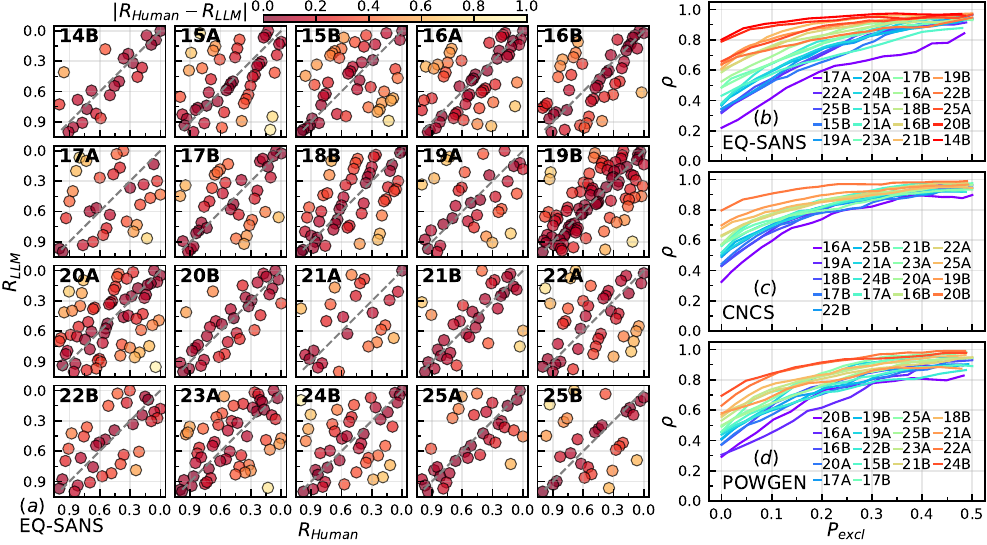}
    \caption{Comparison between Human ranking and LLM ranking. Normalized ranking $R_{LLM}$ and $R_{Human}$ are the ranks divided by total number of proposal in each cycle. (a) $R_{LLM}$ versus $R_{Human}$ for all analyzed run cycles from EQ-SANS. (b) Spearman's $\rho$ correlation versus percentage of excluded outliers for all run cycles at EQ-SANS. (c)-(d) Similar to (b), but for CNCS and POWGEN. }
    \label{fig:rank_3BL}
\end{figure*}

Aggregating these LLM rank $R_{LLM}$ for the past 20 run cycles, and compare with the Human rank $R_{Human}$, Fig.~\ref{fig:rank_3BL}(a) shows the scatter plot of the $R_{LLM}$ versus $R_{Human}$. Within the dataset, we notice some proposals have the same LLM-BT score or human score, to reduce the ambiguity of such situation, we determine the ranking not only based on single key. For the LLM rank, we sort the proposals using BT score as the primary key, the human score as secondary key, and proposal number as tertiary key. Similarly, for the human rank, the human score is the primary key, and the BT score and proposal number are used as secondary and tertiary key, respectively. In this way, proposals obtain the same BT score and human score are ranked in the same order.

Noticeably, all of the scatter plot in Fig.~\ref{fig:rank_3BL}(a) exhibit positive correlation across all run cycles. This positive relationship indicates a strong overall agreement between the rankings produced by the LLMs and the historical rankings determined by human experts. This finding confirms that the underlying quality differences within each run cycle are consistently captured by both assessment methods. Furthermore, the scatter plots provide useful, per-cycle information by visually highlighting outliers proposals where the LLM and human ranks significantly diverge, which can be investigated by the review committee. 

Fig.~\ref{fig:rank_3BL} (b)-(d) show the Spearman's $\rho$ correlation\cite{spearman1961proof,sedgwick2014spearman} between $R_{LLM}$ and $R_{Human}$ for each cycle, versus the portion of the excluded top outliers, for all three beamlines. Spearman's $\rho$ measures the statistical dependence between the rank orderings assigned by the LLM and the human reviewers; a value closer to 1.0 indicates a higher degree of agreement in the relative order of ranked proposals. Intuitively, the correlation increases significantly as more outliers are excluded, and the correlation itself may be used to determine the portion of outlier to be highlighted for further investigation by the reviewer committee. For the full dataset, the correlation distributed around 0.6, as high as 0.8.

\subsection{Publication metric}
\label{ssec:Publication_metric}

Despite the lack of ground truth for determining the proposal ranking, we attempt to evaluate the effectiveness of each ranking results based on the publication record, and investigate the correlation between proposal ranking and corresponding publication productivity. Since rejected proposals do not even have experimental output, we only focus on the accepted proposals in this analysis. We also emphasize that since the past proposal acceptance are based on human ranking, the results is expected to bias towards the human ranking.

By aggregating the proposal dataset with the publication dataset, which include the associated proposal number for each publication, we are able to count the number of publication $N_{pub}$ associated with each proposal. Meanwhile, some publication uses data from more than one proposal, thus we use discounted number of publication $N_{dpub}$ that only count one publication as $1/K$ if it is associated with $K$ proposals.

\begin{figure}[!t]
    \centering
    \includegraphics[width=\linewidth]{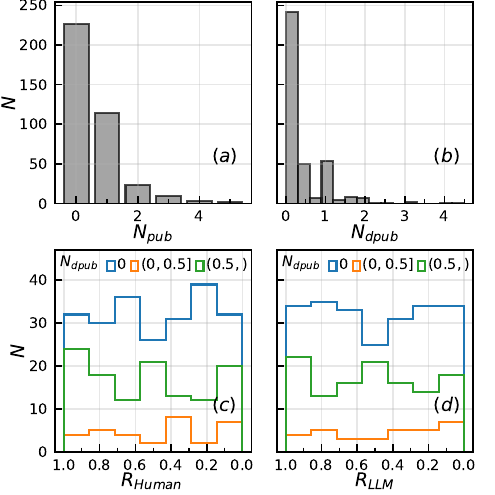}
    \caption{Publication data for all 20 run cycles covered in this work for EQ-SANS. (a) Distribution of the related number of publication $N_{pub}$ for all accepted proposals. (a) Distribution of the discounted number of publication $N_{dpub}$. For each proposal, every published paper with $K$ related proposal is counted as $1/K$. (c) Distribution of normalized human rank $R_{Human}$ for different backset of $N_{dpub}$. (d) Similar to (c) but for LLM rank $R_{LLM}$.}
    \label{fig:publication_rank_hist}
\end{figure}

Fig.~\ref{fig:publication_rank_hist}(a) shows the histogram distribution of the number of proposals $N$ versus the number of publication $N_{pub}$, and Fig.~\ref{fig:publication_rank_hist}(b) shows a similar histogram but with discounted number of publication $N_{dpub}$. Diving deeper into the correlation between the proposal rank and $N_{dpub}$, we plot the distribution of normalized human rank $R_{Human}$ and LLM rank $R_{LLM}$ for proposals divided into three $N_{dpub}$ intervals. If the ranking effectively selects proposals with high publication potential, we would expect a clear separation of distributions in Fig.~\ref{fig:publication_rank_hist}(c) and (d). Specifically, proposals with zero discounted publications ($N_{dpub}=0$) should be concentrated toward lower ranks (normalized rank values closer to 1.0), whereas proposals with high publication output should be concentrated toward higher rank. However, this distinct separation, which would confirm a strong correlation between rank and publication output, is not clearly evident in the distributions shown in Fig.~\ref{fig:publication_rank_hist}(c) and (d).

Furthermore, we calculate the publication metric for each run cycle, defined by:
\begin{equation}
    M_{dpub} = \frac{\sum_{i=1}^{N} (1-R_i)N_{dpub,i}}{\sum_{i=1}^{N} N_{dpub,i}}
\end{equation}
for both human rank $R=R_{Human}$ and LLM rank $R=R_{LLM}$, where the summation $\sum_{i=1}^{N}$ is over all proposals in each run cycle. This metric captures the correlation between the $N_{dpub}$ and normalized rank $R$ for all proposals within a run cycle, and indicate the effectiveness of the ranking for recognizing proposals with higher potential for publication.

\begin{figure}[!t]
    \centering
    \includegraphics[width=\linewidth]{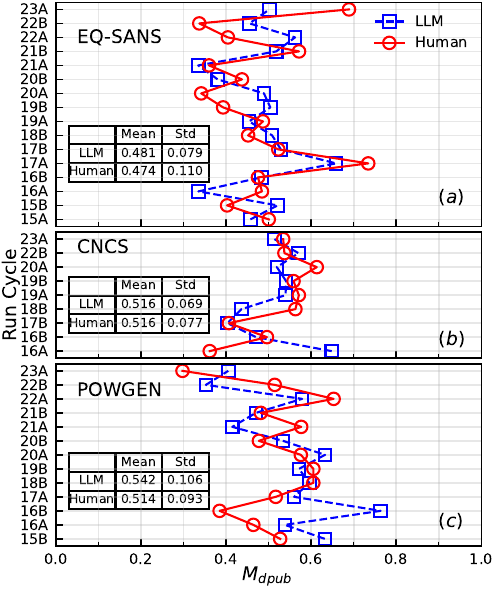}
    \caption{Publication metric $M_{dpub}$ defined by the correlation between rank and discounted number of publication for both $R=R_{Human}$ and $R=R_{LLM}$ versus all run cycles, for (a) EQ-SANS, (b) CNCS, and (c) POWGEN. Only run cycle with at least 4 proposals with associate publication are included.}
    \label{fig:publication_metric}
\end{figure}

The publication metric $M_{dpub}$ for both human and LLM rank over the past run cycles used in current work are shown in Fig.~\ref{fig:publication_metric}. To improve the statistical significance, only run cycles with at least 4 proposals with associate publication are included. Among these run cycle, there is no statistical significant difference observed between the human ranking and LLM ranking. The mean and standard deviation of $M_{dpub}$ over the run cycels are indicated in the figure. For example, for EQ-SANS, With $0.481\pm0.079$ for LLM ranking and $0.474\pm0.110$ for human ranking, it indicate that LLM ranking is no worse than
human ranking in terms of finding proposals with potentials for publication. Similarly, for POWGEN, the metric value ($0.542 \pm 0.106$) is slightly higher than the human metric ($0.514 \pm 0.093$). However, given the standard deviations, this difference is not statistically significant, reinforcing the conclusion that the LLM ranking performs comparably to the human ranking.

\subsection{Cost analysis}
\label{ssec:Cost_analysis}
LLMs intuitively cost significantly less than the human labor for the proposal selection, regardless the quadratic scale of the workload used in the PP approach. To quantify the cost reduction, we carry out estimation of the cost using the labor market data published by the U.S. Bureau of Labor Statistics (BLS), and the token cost from the LLM provider, these numbers are shown in Tab.~\ref{tab:cost_input}. The token usage mean and standard deviation are calculated from all pairwise preference comparison of proposal pairs from all run cycles. The input and output token cost are for the Gemini-2.5-flash model used in this work. For the human data, the salary is given by the BLS 2025 data on the postsecondary engineering teachers, which mostly cover teacher and professor from colleges, universities, and professional schools. The work hour is estimated using the total average weekly working hour in the U.S., adn the review time 1h is a reasonable rough estimation which carries most of the uncertainty, but we will see this won't matter much due to the significance in cost difference.

\begin{table}[!h]
\centering
    \begin{tblr}{
      colspec = {l X[1.2,c] X[c] X[1.2,c] X[c]},
      vlines,
      hline{1,2} = {-}{},      % top and after "LLM"
      hline{3} = {2-5}{},      % after "input / output"
      hline{4} = {-}{},      % after "token / cost"
      %row{1-2} = {font=\bfseries},
      cell{1}{1} = {r=3}{c},   % LLM spans 3 rows, centered
      cell{1}{2} = {c=2}{c},   % input spans 2 columns
      cell{1}{4} = {c=2}{c},   % output spans 2 columns
    }
    LLM     & \SetCell[c=2]{c} input   &   & \SetCell[c=2]{c} output  &   \\
            & token/pair              & cost               & token/pair              & cost \\
            & {$4869 \pm 1081$}       & \$0.3/M            & {$1255 \pm 247$}        & \$2.5/M
    \end{tblr}
    
    %\vspace{1em}
    
    \begin{tblr}{
      colspec = {l X[c] X[c] X[c] X[c]},
      vlines,
      hline{1,3} = {-}{},
      hline{2} = {2-5}{},
      %row{1} = {font=\bfseries},
      cell{1}{1} = {r=2}{c},
    }
    Human   & salary                                      & work hours                          & review time \\
            & 119,340\,\cite{bls2025oes251032}/y         & 41.8\,h/w\,\cite{bls2025cpsaat23}   & $\sim$1\,h/review
    \end{tblr}
\caption{Input for calculating LLM cost and human cost for proposal selection. Token usage statistic are from all LLM runs presented in this work. Human data is from U.S. Bureau of Labor Statistics economic data report.}
\label{tab:cost_input}
\end{table}

\begin{table}[!h]
\centering
    \begin{tblr}{
      colspec = {l X[c] X[c] X[c] X[c]},
      vlines,
      hline{1,2,3} = {-}{},
    }
     cost ratio       & per review    & $N=30$ & $N=70$ \\
     Human/LLMs  & 11,935    & 823   & 346
    \end{tblr}
\caption{Cost ratio between Human and LLMs per review, and for number of proposal $N=30$ and $N=100$, when human use individual scoring and LLMs use pairwise preference}
\label{tab:cost_summary}
\end{table}

Using data in Tab.~\ref{tab:cost_input}, we estimate the human labor cost per proposal review is $\$54.9$, while the LLM cost per proposal pairwise preference is $\$0.0046$. Assuming that, for human to carry out the pairwise preference, the lowered cognition burden cancels with the longer text reading required, and the cost equals. We are able to estimate the human cost for both IS and PP approach and LLM cost for PP approach for different size of the proposal pool in a run cycle. Fig.~\ref{fig:cost_analysis} shows the results. Since the cost for PP scales quadratically with the number of proposal $N$, the LLM cost using the PP will eventually catch up the human cost using IS, but the $N$ will be around 20,000, which is extremely unlikely to happen and also impossible for human to process. For the typical range of number of submitted proposal, $N\in[30,70]$, the human cost is 346 to 823 times of the LLM cost. In other word, for typical proposal review at SNS, the LLM with PP approach cost 0.12\% to 0.29\% of the human reviewers using IS, these results are summarized in Tab.~\ref{tab:cost_summary}. In addition, for very large batch of proposal submission, it is unnecessary to compare every pair of proposal, the comparison matrix can be sparse and the choice of paired proposals can be optimized\cite{jamieson2011active,shah2018simple}. It is also worth emphasizing that the cost of LLMs has been decreasing exponentially over time since the release of GPT-3.5\cite{xiao2025densing}, while the human cost has been increasing over the year\cite{bls_eci_2025}. Therefore the cost gap between human reviewer and LLMs is only going to increase over time.

\begin{figure}[!t]
    \centering
    \includegraphics[width=\linewidth]{cost_analysis.pdf}
    \caption{Estimated total cost versus the number of proposal $N_{pps}$ to rank. Estimation is calculated for LLM uses pairwise preference (PP), and human using both PP and individual scoring (IS).}
    \label{fig:cost_analysis}
\end{figure}

\subsection{LLMs for similarity analysis}
\label{ssec:LLMs_for_similarity_analysis}
Finally, we demonstrate that LLMs can be used to easily evaluate the similarity between any two proposals, which is a task challenging for human reviewers. By passing the proposal text into the embedding model from the LLMs, we can transform each 2-page proposal into a high dimensional vector $\vb{V}$. These embedding vectors $\{\vb{V}_i\}$ reside in a continuous, high-dimensional semantic space where geometric proximity directly reflects the degree of meaning-level similarity between the underlying texts. Because modern embedding, e.g. Qwen3-embedding-8b\cite{bai2023qwen} with a 4096 dimensional embedding space, are trained on massive corpora to predict contextual relationships, proposals that share similar scientific goals, methodologies, or even stylistic patterns are mapped to nearby points in this space. This property enables robust, automated similarity analysis: two proposals can be quantitatively compared almost instantly by simply computing the similarity score of their embedding vectors $\frac{\vb{V}_i\cdot\vb{V}_j}{|\vb{V}_i||\vb{V}_j|}$, without requiring the LLMs to generate any text. In contrast, human reviewers often struggle to maintain consistent similarity judgments across large proposal sets due to cognitive load, fatigue, and subjective bias.

Beyond the accuracy and consistency, the LLM embedding approach offers dramatic efficiency advantages. Computing embeddings for $N$ proposals requires only a single forward pass per documents, thus a $O(N)$ complexity, while the all-pairs similarity matrix is obtained instantly via highly optimized batch dot-product operations, making the quadratic term effectively negligible. In contrast, human reviewers must perform a genuine $O(N^2)$ effort, reading and comparing nearly every pair. This embedding method thus transforms an inherently quadratic human burden into a fast, linear-time operation, enabling routine large-scale similarity analysis and duplicate detection.

\begin{figure}[!t]
    \centering
    \includegraphics[width=\linewidth]{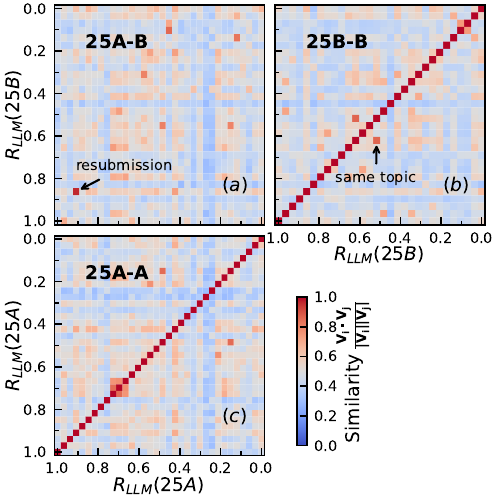}
    \caption{Similarity analysis for EQ-SANS' run cycle 25A and 25B using the LLM embedding model. Cosine similarity between two embedding vector $\vb{V}_i$ and $\vb{V}_j$ is $\frac{\vb{V}_i\cdot\vb{V}_j}{|\vb{V}_i||\vb{V}_j|}$. (a) Inter-cycle similarity between run cycle 25A and 25B. (b) Intra-cycle similarity within the 25B. (c) Intra-cycle similarity within the 25A.}
    \label{fig:similarity_analysis}
\end{figure}

Fig.~\ref{fig:similarity_analysis} shows examples of the similarity matrix based on the proposals from the 25A and 25B cycles. These heatmaps not only provide a straightforward analysis on the overall intra-cycle and inter-cycle similarity between proposals, but also become very useful for quick filtering for further investigation. For example, Fig.~\ref{fig:similarity_analysis}(a) shows the inter-cycle similarity between the proposals from the 25A and 25B cycle, the highest similarity point, as highlighted in the plot, is actually a resubmission of a revised proposal. Moreover, in Fig.~\ref{fig:similarity_analysis}(b), the highest similarity point turns out to be a result from two proposals on the same topic, even though they are submitted from different principal investigators. Based on this similarity value, one can manually determine a threshold value for further investigation, and pass the highlighted pairs in to LLMs to summarize the similarity and difference between these proposals before passing to human expert for further verification.

\section{Summary}

%1 what have we done
In this work, we demonstrate LLMs' potential for proposal selection at large user facilities. Without loss of generality, we used the historical proposal data from the three most representative beamlines in SNS at ORNL. We use LLMs to carry out pairwise preference analysis between every pair of proposals within the same cycle, and calculate the rank based on the win-lose results. The LLM rank $R_{LLM}$ is positively correlated with the human rank $R_{Human}$, with Spearman's $\rho$ correlation varying from 0.2 to 0.8. To evaluate the effectiveness of the ranking, we study the correlation between publication output from the proposals and the corresponding $R_{LLM}$ and $R_{Human}$. We found no statistical difference in the effectiveness of the LLM ranking compared to the human ranking in identifying proposals with high publication potential. We then analyze the dollar cost for obtaining the human and LLM ranking, and estimate that the LLMs approach is $1/823$ to $1/346$ of the human approach for reasonable size of the proposal pool. Finally, we demonstrate that the embedding model from the LLMs can be used to perform the similarity analysis between any two proposal with linear algorithmic complexity, comparing to human's quadratic complexity, while providing reliable quantitative value.

%2. what is the significance
The fundamental objective of proposal selection is to establish the relative strengths among all submitted proposals and generate a highly reliable rank. Given this goal, the LLM-enabled pairwise preference approach stands out as a logically superior and more robust methodology. Our study not only demonstrates that the LLM ranking system performs comparably to human reviewers but also reveals that the cost is effectively negligible for any large user facility. Needless to say, the overall framework provided in this work is not limited to the specific user facility, and can be easily applied to other facilities at ORNL, other national laboratories, and even major funding agencies like DOE, NSF and NIH. This LLM-assisted proposal selection system can significantly augment the capabilities of scientific review committees by delivering essential, high-quality ranking information at a negligible cost. 

%3 what are the future directions
Looking forward, we see two main directions for further utilizing and enhancing this LLM framework for proposal selection. One obvious direction is increasing the scale of the experiments, scale to more facilities, and test different LLMs. Another direction with more practical impact is fine-tuning an LLM using the past proposal and publication data to predict the possibility for publication for a submitted proposal, which has the potential to significantly boost the productivity of large user facilities.

\section*{Data availability}
The scripts for this work are available at the GitHub repository \url{https://github.com/ljding94/ProposalArena}. We are unable to provide the proposal text analyzed in this work.

\begin{acknowledgments}
We thank Jeff Patton for gathering and consolidating all proposals and science reviews for this study. This research used resources at the Spallation Neutron Source, a DOE Office of Science User Facility operated by the Oak Ridge National Laboratory.
\end{acknowledgments}

%\newpage
%\nocite{*}

\section*{Reference}
\bibliography{reference}% Produces the bibliography via BibTeX.

\end{document}